\documentclass[journal,twoside,web]{ieeecolor}
\usepackage{generic}
\usepackage{cite}
\usepackage{amsmath,amssymb,amsfonts}
\usepackage{graphicx}
\usepackage{textcomp}

\usepackage[colorlinks,urlcolor=blue,linkcolor=blue,citecolor=blue]{hyperref}
\usepackage{booktabs}
\usepackage{longtable}
\usepackage{multirow}
\usepackage{makecell}
\usepackage{caption}
\usepackage{tikz}
\newcommand*{\circled}[1]{\lower.7ex\hbox{\tikz\draw (0pt, 0pt)%
circle (.5em) node {\makebox[1em][c]{\small #1}};}}
\usepackage[noend]{algpseudocode}
\usepackage{algorithmicx,algorithm}

\def\BibTeX{{\rm B\kern-.05em{\sc i\kern-.025em b}\kern-.08em
    T\kern-.1667em\lower.7ex\hbox{E}\kern-.125emX}}
\begin{document}
\title{Temporal-spatial Correlation Attention Network for Clinical Data Analysis in Intensive Care Unit}
\author{
\IEEEmembership{Weizhi Nie}, 
\IEEEmembership{Yuhe Yu}, 
\IEEEmembership{Chen Zhang},
\IEEEmembership{Dan Song*}, 
Lina Zhao, 
Bai Yunpeng 
\thanks{Manuscript received April 19, 2005; revised August 26, 2015. This work was supported in part by the National Natural Science Foundation of China (62372337) and the Natural Science Foundation of Tianjin (16JCZDJC31100, 16JCZDJC31100).}
\thanks{Weizhi~Nie, Yuhe~Yu, Chen~Zhang, Dan~Song are with the school of electrical and information engineering, Tianjin University (e-mail: weizhinie@tju.edu.cn; yuyuhe@tju.edu.cn; zhangchen001@tju.edu.cn; dan.song@tju.edu.cn).}
\thanks{Lina~Zhao, is with Department of Anesthesiology, Tianjin Medical University General Hospital.}
\thanks{Bai~Yunpeng, MD  Postdoctor, is with Department of Cardiac surgery, Tianjin University Chest Hospital (e-mail: oliverwhite@126.com).}
}

\maketitle

\begin{abstract}
In recent years, medical information technology has made it possible for electronic health record (EHR) to store fairly complete clinical data. This has brought health care into the era of ``big data''. 
However, medical data are often sparse and strongly correlated, which means that medical problems cannot be solved effectively. With the rapid development of deep learning in recent years, it has provided opportunities for the use of big data in healthcare. In this paper, we propose a temporal-saptial correlation attention network (TSCAN) to handle some clinical characteristic prediction problems, such as predicting death, predicting length of stay, detecting physiologic decline, and classifying phenotypes. Based on the design of the attention mechanism model, our approach can effectively remove irrelevant items in clinical data and irrelevant nodes in time according to different tasks, so as to obtain more accurate prediction results. Our method can also find key clinical indicators of important outcomes that can be used to improve treatment options. Our experiments use information from the Medical Information Mart for Intensive Care (MIMIC-IV) database, which is open to the public. Finally, we have achieved significant performance benefits of 2.0\% (metric) compared to other SOTA prediction methods. We achieved a staggering 90.7\% on mortality rate, 45.1\% on length of stay. The source code can be find: \url{https://github.com/yuyuheintju/TSCAN}.
\end{abstract}


\section{Introduction}
\label{sec:introduction}
\IEEEPARstart{T}{he} explosive growth of electronic healthcare data and the rapid development of artificial intelligence technology have facilitated the deep application of healthcare data in multiple scenarios~\cite{DBLP:journals/titb/ShickelTBR18}. On the one hand, the rapid development of the Internet of Things (loT) and big data technologies provides technical support for the large-scale acquisition, safe storage, and rapid analysis of medical health data. The health care sector has now entered the era of big data. On the other hand, artificial intelligence has reached a new stage, with both theory and technology making big steps forward. Machine learning, especially deep learning, provides technical support for intelligent medical applications.

Electronic Health Record (EHR) contains a variety of medical time-series data organized in different forms, such as records of patients' admission times, medications taken and procedures scheduled, observations of laboratory results, etc. The health status of patients in Intensive Care Unit (ICU) needs to be monitored in real time to ensure that the health care staff can provide timely and effective care and that the treatment plan can be adjusted according to the development of the patients’ condition. Therefore, indicators of a patient's physical health that are regularly monitored, such as respiratory rate, heart rate, and blood pressure, often contain a wealth of information that can be used to formulate a patient's treatment strategy.

Quantifying patient health and predicting future outcomes is an important area of critical care research~\cite{DBLP:conf/mlhc/JohnsonPM17}. One of the most immediately relevant outcomes for the ICU is in-hospital mortality, leading many studies toward the development of mortality prediction models.
The accuracy of early mortality prediction can benefit the formulation of patients’ care strategies. 
Traditionally, discharge date estimation is done manually by clinicians, which is time-delayed and unreliable~\cite{DBLP:journals/corr/abs-2007-09483}. Deep learning models based on EHR can improve the accuracy of length of stay prediction, which can reduce the burden on clinicians in terms of administration. Meanwhile, decompensation prediction enables more complex decision planning, and phenotype classification can also be used early on as a complementary aid to the doctor's diagnosis.

However, the value of medical time series is huge but the density is very low. These situations have brought great challenges to accurate data analysis. As a result, the core problem of health care big data research is how to use artificial intelligence technology to fully explore the value in data and realize the transformation of data values into applications. Most of the work being done now uses a small number of physiological variables and doesn't pay much attention to correlations of different variables. For instance,~\cite{DBLP:journals/corr/HarutyunyanKKG17,DBLP:journals/jbi/PurushothamMCL18} proposed benchmarks to address four meaningful clinical tasks.~\cite{DBLP:journals/corr/LiptonKEW15} proposed to use LSTM recurrent neural networks to handle clinical diagnosis.~\cite{DBLP:journals/corr/abs-2007-09483} proposed a temporal pointwise convolutional networks to handle length-of-stay prediction. However, all of these methods often lose too much relevant diagnostic data due to the sparsity of the data. At the same time, they can't say how important the diagnostic information is for the task, and it's not easy for them to give a guiding diagnosis and treatment plan.

We propose a temporal-spatial correlation attention network (TSCAN) to address various clinical characteristic prediction problems, such as predicting death, predicting length of stay, detecting physiologic decline, and classifying phenotypes. Our proposed temporal and spatial attention networks are designed to facilitate multiple core information fusion and extraction tasks. To fully exploit the information in the temporal dimension of the medical time series, we introduce a recursive and merged attention mechanism, which enables the fusion of information from previous time periods in the current time period and allows for an orderly exploration of the hidden information in the temporal dimension when dealing with long time series. Additionally, we consider both temporal and spatial dimensions, taking into account the relevance of both dimensions and fusing them, making our approach more comprehensive than most studies on medical time series, which only consider temporal connections.

Our method can effectively remove irrelevant items in clinical data as well as irrelevant nodes in time based on different tasks, resulting in more accurate prediction results. Our method can also find key clinical indicators of important outcomes that can be used to improve treatment options. Finally, we select data from the MIMIC-IV dataset to demonstrate the performance of our approach.
 
The contributions of this paper are as follows:
\begin{itemize}
\item We propose a temporal-spatial correlation attention network (TSCAN) to handle some clinical characteristic prediction problems, All indicators have exceeded the current optimal algorithm;
\item We expanded the physiological variables in medical time series to 155 with the advice of professionals in collaboration with local hospitals. We also publish these data\footnote{https://github.com/yuyuheintju/TSCAN} for other researchers' work;  
\item We discuss the contribution ratios of the 155 physiological variables selected for the task on the advice of professionals, which have important implications for subsequent optimization of the model and medical recommendations.
\end{itemize}

The remainder of this article is organized as follows. Section 2 illustrates the process of data selection. Section 3 presents several related works. Section 4 provides the details of our approach. The corresponding experimental results and analysis are given in Section 5. Finally, we discuss the limitations and our future work and conclude this paper in Section 6.

\section{Related Work}
EHR contains different types of medical time-series data, and a great deal of research has been accomplished in medical time series analysis. The key to the research is how to automatically extract temporal associations and long-range dependencies from the data, and the main methods of the research focus on traditional machine learning methods and deep learning methods based on neural networks, of which deep learning-based medical time series analysis has gradually become a major research direction in recent years.

\subsection{Traditional machine learning algorithms}
Traditional machine learning methods for building predictive models based on medical time-series data have been developed over decades, including conditional random field (CRF)~\cite{10.1007/11569541_47}, Expectation-Maximization algorithm (EM)~\cite{friedman2013bayesian}, Bayesian networks~\cite{van2014learning} and others. Caruana et al.~\cite{10.5555/2998828.2998963} showed that backpropagation can be effective in medical data analysis, and Cooper et al.~\cite{cooper1997evaluation} demonstrated the application of eight machine learning algorithms, including K-nearest neighbor (KNN), in predicting patients' survival. Awad et al.~\cite{awad2017early} used a random forest algorithm to predict early mortality of patients, and Bayesian networks~\cite{lucas2004bayesian,sierra2001using} were used to develop multi-label classifiers, which were applied to medical multi-classification problems. Logistic Regression (LR) was widely used in the field of predicting mortality risk among hospitalized patients~\cite{rosenberg2002recent,knaus2002apache,le1984simplified,cooper1997evaluation} and was chosen as one of the baselines.

\subsection{Deep learning algorithms}
In recent years, deep learning technology with deep neural networks as the core has been gradually emerging in the field of medical applications with the increase of computers' computing power. Lasko et al.~\cite{lasko2013computational} used hierarchical autoencoders (AEs) based on time-series data of uric acid measurements to predict the risk of gout and acute leukaemia. Chen et al.~\cite{DBLP:conf/sdm/ChengWZH16} proposed a model based on CNN to extract local temporal associations in time-series data for the prediction of futural risk of congestive heart failure (CHF) and chronic obstructive pulmonary disease (COPD). Mobley et al.~\cite{mobley1995artificial} applied ANN to length of stay prediction, Grigsby et al.~\cite{grigsby1994simulated} considered length of stay prediction as a binary classification task in order to identify patients at risk of long-term hospitalisation in early stage, and Emma et al.~\cite{DBLP:journals/corr/abs-2007-09483} proposed the Temporal Pointwise Convolution (TPC) achieving relatively good results in this task. 

Lipton et al.~\cite{DBLP:journals/corr/LiptonKEW15} extracted 13 variables of laboratory outcome from paediatric ICU patients and used Long Short-Term Memory networks (LSTM) to learn temporal associations and long-distance dependencies in time-series data. Choi et al.~\cite{DBLP:journals/jamia/ChoiSSS17} used a variety of RNN models such as Gated Recurrent Unit (GRU) and LSTM networks to process clinical time-series data for the prediction of medical events. It is worth mentioning the extensive application of LSTM in clinical prediction tasks, including prediction of cardiac arrest~\cite{DBLP:conf/mlhc/TonekaboniMLEGG18}, acute kidney injury~\cite{tomavsev2019clinically}, missing data inference~\cite{cao2018brits}, prediction of drugs~\cite{DBLP:journals/corr/ChoiBS15,DBLP:conf/mlhc/SureshHJCSG17} and length-of-stay prediction~\cite{DBLP:journals/corr/abs-1910-00964}.

\subsection{Application on specific tasks}
In the task of mortality prediction,~\cite{clermont2001predicting,celi2012database} illustrated that artificial neural networks (ANN) can achieve better results than Logistic Regression. For decompensation prediction tasks, the Recurrent Attentive and Intensive Model (RAIM)~\cite{10.1145/3219819.3220051} and Simply Attend and Diagnose (SAnD)~\cite{song2018attend} have applied attention mechanisms to predict physiological decompensation of critical patients in ICU. Phenotype classification is also a promising task and temporal convolutional networks~\cite{DBLP:journals/corr/RazavianMS16} and feedforward networks~\cite{DBLP:conf/kdd/CheKLBL15,lasko2013computational} have already be used in predictions of diagnostic codes based on medical time-series data. There were researches indicating that deep learning models represented by LSTM performs well in predicting mortality~\cite{DBLP:journals/corr/ChePCSL16,shickel2019deepsofa}, length of stay~\cite{DBLP:journals/corr/abs-1910-00964} and diagnostic classification~\cite{rajkomar2018scalable}. So LSTM~\cite{DBLP:journals/corr/HarutyunyanKKG17} was chosen as another baseline.

\begin{figure*}[ht!]
\centering
\includegraphics[width=0.8\textwidth]{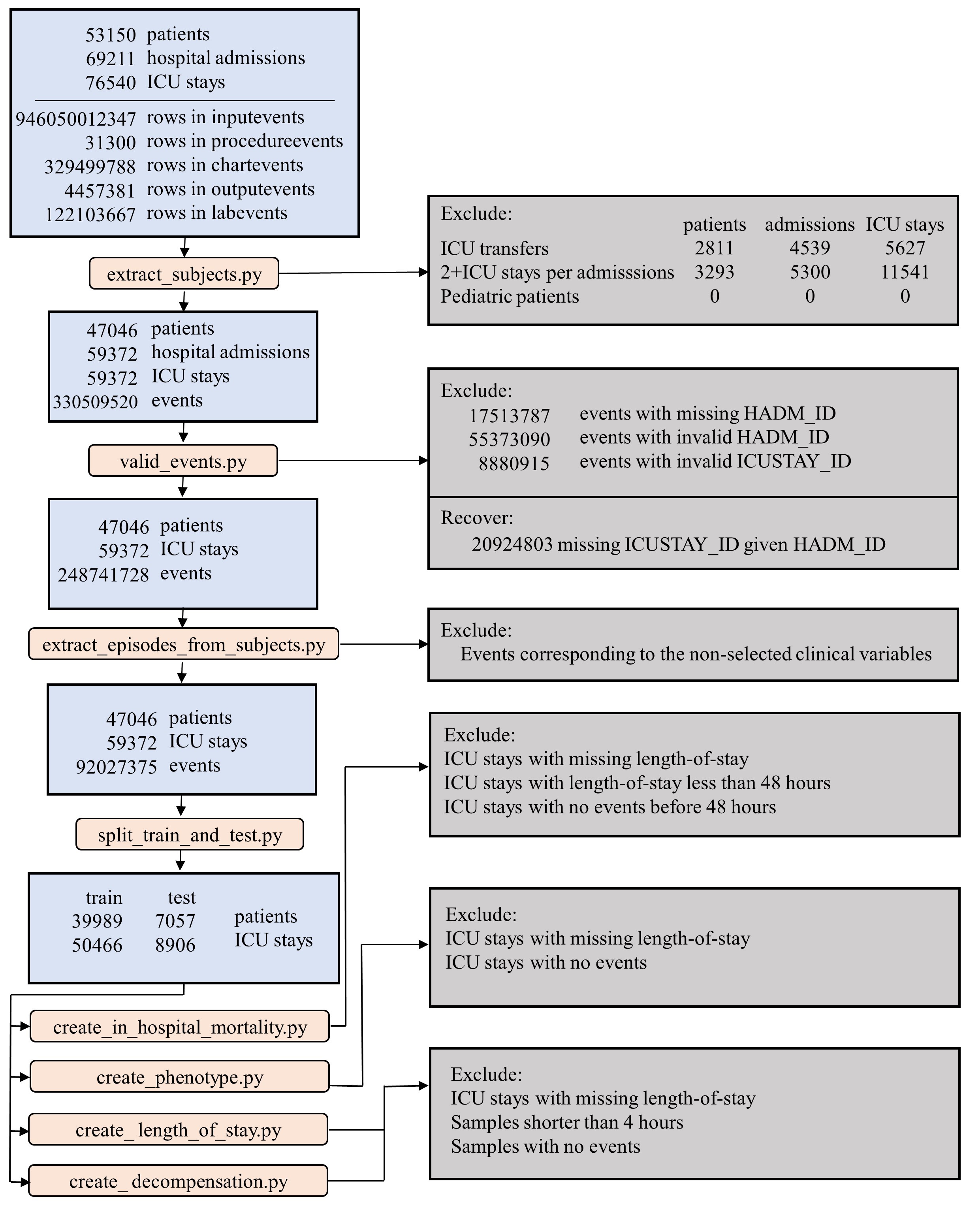}
\caption{\label{fig:processess} Data selection. The critical care database of MIMIC-IV contains 76540 ICU stays from 53150 patients admitted, and after four data processing scripts, we can obtain 47046 ICU stays from 59372 patients admitted. Then further filtered according to the requirements of the different tasks.}
\end{figure*}

\section{MIMIC-IV Dataset}
\begin{table*}[!t]
\caption{The 24 selected clinical variables.  155 physiological variables were selected on the advice of the local hospitals during the collaboration, including 5 categorical variables and 150 continuous-value variables. Here are shown 24 of them.} 
\label{tab:24_variables}
\centering
\begin{tabular*}{0.7\hsize}{@{\extracolsep{\fill}}l c c }
\toprule
Variable   & MIMIC-IV table     & Model as    \\
\midrule     
Albumin                            & chartevents           & continuous  \\
Anion gap                          & chartevents           & continuous  \\
Capillary refill rate              & chartevents           & categorical \\
Cholesterol                        & chartevents           & continuous  \\
Diastolic blood pressure           & chartevents           & continuous  \\
Fraction inspired oxygen           & chartevents           & continuous  \\
Glascow coma scale eye opening     & chartevents           & categorical \\
Glascow coma scale motor response  & chartevents           & categorical \\
Glascow coma scale total           & chartevents           & categorical \\
Glascow coma scale verbal response & chartevents           & categorical \\
Glucose                            & chartevents,labevents & continuous  \\
Heart Rate                         & chartevents           & continuous  \\
Height                             & chartevents           & continuous  \\
Hemoglobin                         & chartevents           & continuous  \\
Magnesium                          & chartevents           & continuous  \\
Mean blood pressure                & chartevents           & continuous  \\
Oxygen saturation                  & chartevents,labevents & continuous  \\
Prothrombin time                   & chartevents           & continuous  \\
Respiratory rate                   & chartevents           & continuous  \\
Systolic blood pressure            & chartevents           & continuous  \\
Temperature                        & chartevents           & continuous  \\
Troponin-T                         & chartevents           & continuous  \\
Weight                             & chartevents           & continuous  \\
pH                                 & chartevents,labevents & continuous  \\
\bottomrule
\end{tabular*}
\end{table*}

\subsection{Data Description}
We use the Medical Information Mart for Intensive Care (MIMIC-IV v0.4) dataset~\cite{johnson2020mimic}. The MIMIC-IV database is a public database of clinical data that researchers from all over the world can use for free. The database has clinical information on more than 380,000 patients who were admitted to Beth Israel Deaconess Medical Center in Boston, Massachusetts, USA, from 2008 to 2019. Like other EHR databases, it keeps detailed information on patients' demographics, lab tests, medication administration, vital signs, surgical operations, disease diagnosis, medication management, survival status, and more. 

MIMIC-IV uses a modular approach to organize data. It has three modules that are made up of 27 tables that make it easy to use data from different sources separately or together. Another advantage is the protection of patients' privacy, which is de-identified in two steps: the first is to replace the patient identifier, hospital identifier, and so on with a random code, and the second is to add a random number of days to the date data fixed for each patient.

\subsection{Data Selection}
The data pre-processing workflow is illustrated in Fig.\ref{fig:processess}. The MIMIC-IV critical care database contains 76540 ICU stays from 53150 patients admitted.

In the first step, the important data are taken straight from the original MIMIC-IV tables and put in order by subject, which is another word for patient. This step has two parts: keeping only adults who were in the ICU (over the age of 18), who were in the ICU only once, and who did not move between ICU wards or wards during the same hospital stay. In this step, differences in how children's and adults' bodies operate and other unclear factors are taken out. This leaves 47,046 unique patients with a total of 59,372 ICU stays.

In the second step, we exclude clinical events that cannot be matched with ICU stays. Firstly, it considers events missing admission ID (HADM\_ID), and only events with HADM\_ID are reserved. Secondly, since the $stays.csv$ table allows the connection of HADM\_ID with ICU stays, we exclude events that do not have HADM\_ID in $stays.csv$. Then, for events where ICU stay IDs (ICUSTAY\_ID) are defaults, they can be recovered by attempting to check the ID. Finally, we only retain events for which the ICUSTAY\_ID is present in $stays.csv$.

In the third step, a patient has a single ICU admission, which is also described as an ``episode''. In this script, for each episode, a list of events is put together in order of time, with only the variables from a list that has already been set. We use 155 physiologic variables, expanding from 17 variables, which are a subset from the Physionet/CinC Challenge 2012~\cite{silva2012predicting}. During the collaboration, the local hospitals helped choose the 155 physiological variables, which included 5 categorical variables and 150 continuous value variables. The 24 variables of 155 are list in Table.\ref{tab:24_variables}. Thus far, we have obtained over 92 million events from five tables (icu/inputevents, icu/procedureevents, icu/chartevents, icu/outputevents, hosp/labevents). Finally, we fixed a test set of 15\% (7,057) of patients, including 8,906 ICU stays.

\section{Our Approach}
\begin{figure*}[ht!]
\centering
\includegraphics[width=\textwidth]{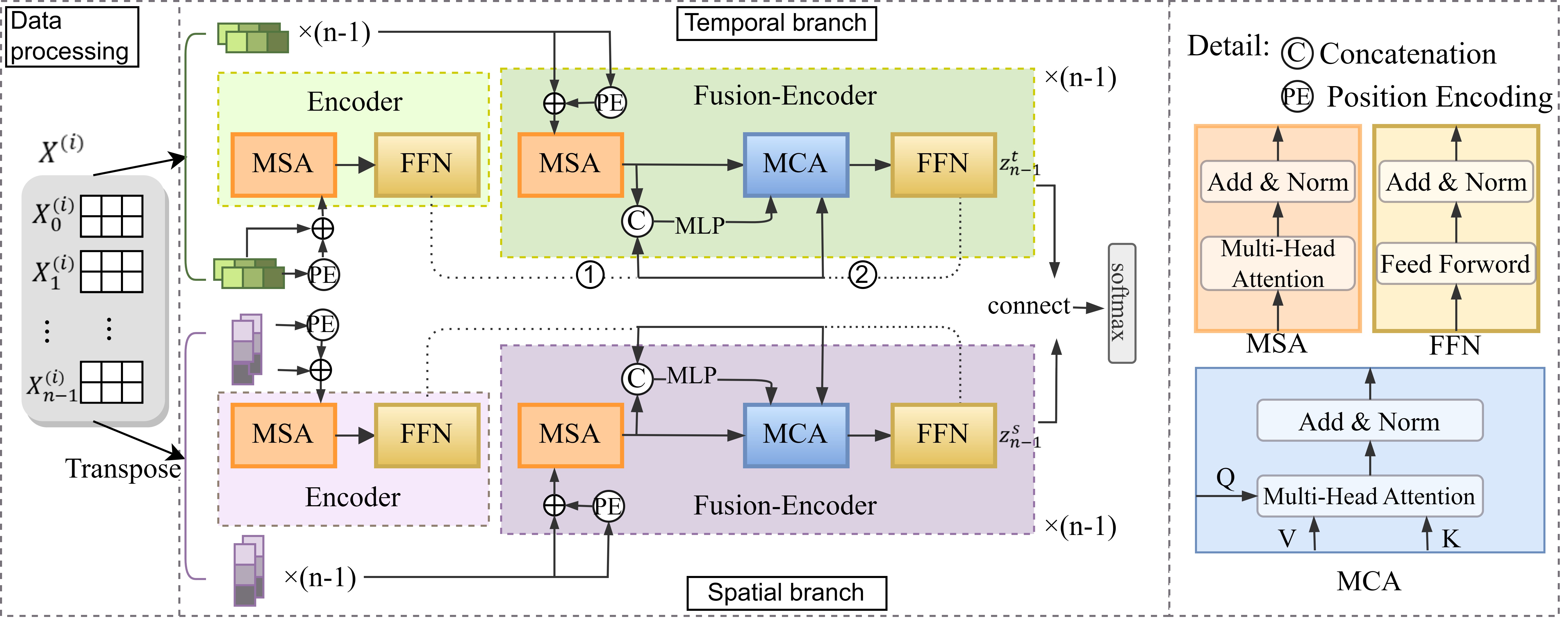}
\caption{\label{fig:model} TSCAN. It can be divided into two branches in both space and time, which is the same tandem style with one Encoder and and $(n-1)$ Fusion-Encoder. In the temporal domain, the input is divided equally into $n$ parts into the model and an additional step of input transposition allows to utilize the attention in the characteristic field for spatial domain.}
\end{figure*}

\begin{figure}[ht!]
\centering
\includegraphics[width=0.45\textwidth]{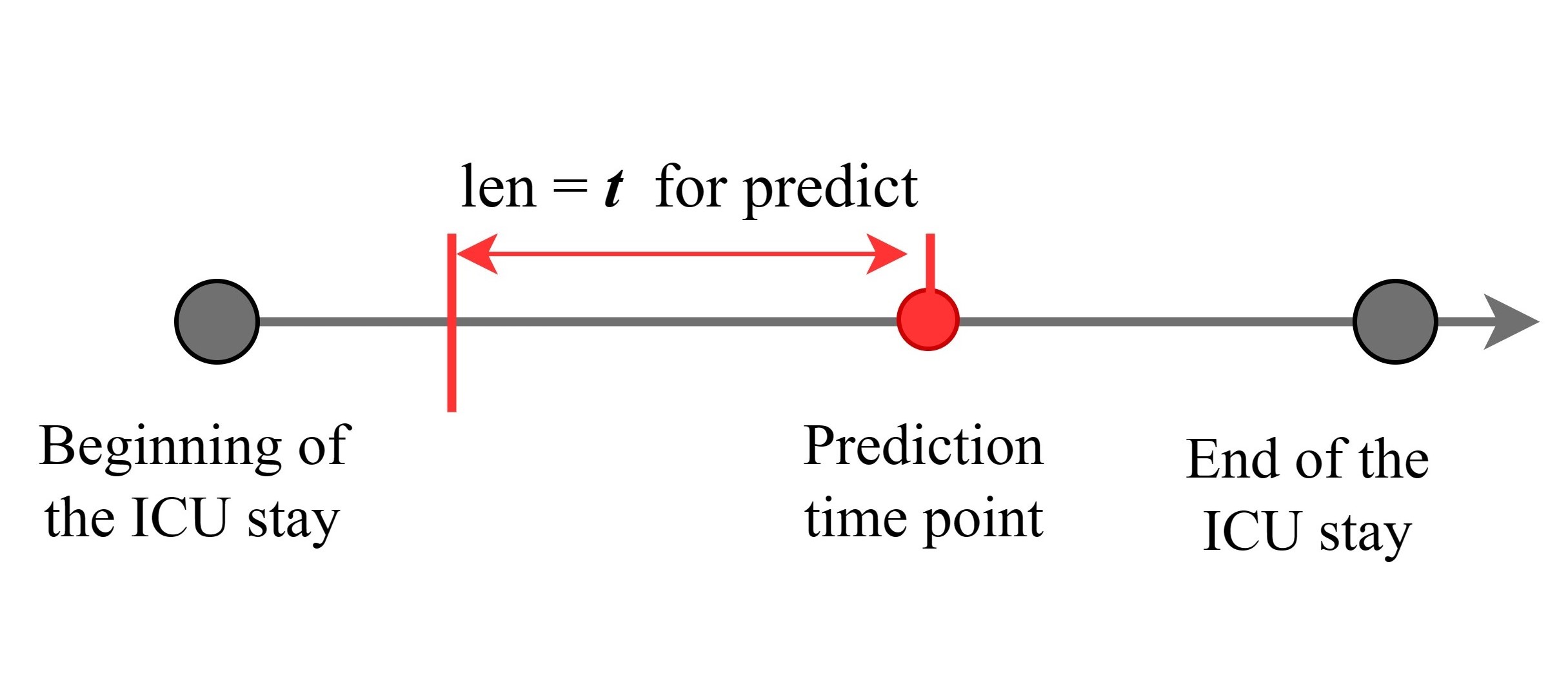}
\caption{\label{fig:overview} Problem define. We choose the data of $t$ hours before the prediction time point for prediction tasks.}
\end{figure}

Our model is designed to find correlations in both time and space, taking into account both temporal trends and connections between features. We hope to integrate the features of the previous time period when processing information of the end time period for long medical time sequences, and to fully exploit the hidden features of the time dimension. For the task of predicting in-hospital deaths, we want to know how much each of the 155 variables we chose in advance has to do with it. The model needs to figure out which of the three indicators—blood protein, blood pressure, and heart rate—is the best predictor of hospital mortality. This will help doctors focus on the most important information about their patients. The framework is illustrated in Fig.\ref{fig:model}, which includes two key modules: 1) Encoder: It is used to obtain the feature vector of each element in both temporal and spatial dimension; 2) Fusion-Encoder: It is used to fuse the information of these elements and obtain the final feature at the end time. The module enables a time-segmented solution to integrate information from all time periods and takes advantage of the features of the time dimension.
\subsection{Problem Define}

For each patient, we need to predict in-hospital mortality, and other tasks in each time interval, such as the length-of-stay prediction. Fig.\ref{fig:overview} shows the problems. To predict, we only use previous clinical data from each time interval. As shown in Fig.\ref{fig:overview}, we chose the data from $t$ hours before the prediction time point as the reference data for the predicting related tasks in this paper. 

For each patient $i$, we can obtain a set of samples $X^i=\{x^i_1,x^i_2,....,x^i_{t}\}$, where $X^i\in R^{t\times d}$. $t$ denotes that we select $t$ times clinical data before prediction time point. $d$ denotes that the $d$ clinical test values are selected in each hour. In Section 3, we select $155$ clinical values based on the advice of Professor Doctor. Here, we apply the one-hot method \cite{RODRIGUEZ201821} to handle clinical data and convert these clinical values in to the format of vector. Finally, in each hour, clinical data $1\times 155$ can be convert to $1\times d$ dimension vector for final predict task. 

In the prediction task, our model needs to map medical features to clinically meaningful labels such as length of stay, in-hospital mortality, etc. A predictive model for c classification can be represented as a mapping from input to category, i.e.$f(X^i): R^{t \times d} \rightarrow\{0,1, \ldots, c\}$ and the entire sample data set is then denoted as $D=\{X^i, y^i\}_{i=1}^N$, where $y_i$ is the target label, $N$ is the number of samples. For the in-hospital mortality task, the patient target label is $y\in\{0,1\}$, where $0$ denotes survival and $1$ denotes death. The predictive model is derived from the input data $X^i$ to obtain the output $y^*_i=f(X^i, \theta)$, where $\theta$ is the parameters of the model. The key of our work is to obtain the best parameter $\theta^{*}$ for different tasks, which can minimize the difference between the output of model $y^*_i$ and the true label $y_i$. In the next subsection, we will detail the structure of our model. 

\subsection{Temporal-spatial Correlation Attention Network}
In order to consider the temporal and spatial information, we apply the two-branch network as Fig.2. Each branch has the same structure, which includes the initial Encoder $E_e$ and the Fusion-Encoder $E_f$. 

We divided each sample $X^i$ into $n$ equal parts based on the time dimension and the model follows these chunks for input: 
\begin{equation}
\begin{aligned}
X^i=[X^i_0,X^i_1,...,X^i_{n-1}],~~~X^i_j \in R^{\frac{t}{n}\times d}.
\end{aligned} 
\end{equation}

\textbf{Temporal Branch.} As shown in the upper part of Fig.\ref{fig:model}, this branch consists of two parts: an Encoder and $n-1$ Fusion-Encoder. $X_0^i$ is the input of Encoder and the output is denoted as $z_0$:
\begin{equation}
z_0=E_e(X_0^i;\omega).
\end{equation}
where $\omega$ denotes the network parameters of Encoder. Encoder's core structure is based on a multi-head self-attention mechanism. The calculation in Encoder can be expressed in detail as follows:
\begin{equation}
\begin{aligned}
&E_{1}= PE(X_{0}^{i}) \\
&z_{0}^{\prime} = MSA(X_{0}^{i}+E_{1}) \\
&z_{0}=FFN\left(z_{0}^{\prime}\right),
\end{aligned}
\end{equation}

Then we process the following set of data, $X^i_j,j=1,...,n-1$, and merge it with previous data information, $z_{j-1}$. The Fusion-Encoder $E_f$ has the same internal structure, but there is a recursive relationship. The j-th Fusion-Encoder is represented by:
\begin{equation}
\begin{aligned}
z_{j}=E_f(X_{j}^{i}, z_{j-1};\delta_j),~~~j =1, 2,..., n-1.
\end{aligned}
\end{equation}
where $\delta_j$ denotes the network parameters of the j-th Fusion-Encoder. When $j=1$, connect line \circled{1} in model and the second input of the 1st Fusion-Encoder is the output of Encoder, $z_0$. For $j=2, 3 ,..., n-1$, the second input of Fusion-Encoder is the output of the last Fusion-Encoder, $z_{(j-1)}$, at which point line \circled{2} is connected. The calculation in the j-th Fusion-Encoder is as follows:
\begin{equation}
\begin{aligned}
&E_{j}  = PE(X_{j}^{i}) \\
&Q = MSA(X_{j}^{i}+E_{j}) \\
&K = z_{j-1} \\
&V = MLP(\text { Concate }(Q, K)) \\
&z_{j}^{\prime}  = MCA(Q, K, V) \\
&z_{j}  = FFN(z_{j}^{\prime})
\end{aligned}
\end{equation}

The core is a multi-head cross-attention mechanism in which Q is derived from  self-attentive output of the j-th time series $X^i_j$, K from the previous Fusion-Encoder's output, and V from a mapping that combines Q and K. In this case, V focuses on a relatively global feature, so the recursive structure can also focus on global information.

\textbf{Spatial Branch.} We transpose time series blocks of $n$ equal parts as the input of the spatial branch to obtain the relevance of variable dimensions by attention mechanism. As shown in Fig.\ref{fig:model}, {\small $(X_{j}^{i})^{T} \in R^{d \times \frac{t}{n}}$}, replaces $X_j^{i}$ of temporal branch. There is no additional difference between this branch and the upper one, both are composed of Encoder and Fusion-Encoder.

Finally, we can obtain $z_{n-1}^t$ from the temporal branch and $z_{n-1}^s$ from the spatial branch. Here, we connect $z_{n-1}^t$ and $z_{n-1}^s$ as the final feature of patient in the predict time point, and apply the Softmax to train and obtain final predict result. Algorithm 1 shows the overall training process. 

\begin{algorithm}[t]
\caption{Temporal-spatial Correlation Attention} 
\hspace*{0.02in} {\textbf{Require:}} 
the initialization parameters $\omega^t_{0},\delta^t_{0},\omega^s_{0},\delta^s_{0},$ the number of epochs $M,$ the number of batches in each epoch $I$ and
the number of parts $n$.
\begin{algorithmic}[1]
\State $\omega^t\leftarrow\omega^t_{0},~\delta^t\leftarrow\delta^t_{0},~\omega^s\leftarrow\omega^s_{0},~\delta^s\leftarrow\delta^s_{0}$~(Initialize
parameters)
\State $m\leftarrow0,~i\leftarrow0$~(Initialize the counter)
\While{$m<M$} 
\State $m\leftarrow m+1$
\While{$i<I$} 
\State $i\leftarrow i+1$
\For{Temporal branch} 
\State $z_0^t=E_e(X_0^i;\omega^t)$
\For{$j\leftarrow1~\textbf{to}~n-1$}
    \State $z^t_{j}=E_f(X_{j}^{i},z^t_{j-1};\delta^t_j)$
\EndFor
\EndFor
\For{Spatial branch}
\State $z_0^s=E_e(X_0^i;\omega^s)$
\For{$j\leftarrow1~\textbf{to}~n-1$}
    \State $z^s_{j}=E_f((X_{j}^{i})_T,z^s_{j-1};\delta^s_j)$
\EndFor
\EndFor
\State Concate $z_{n-1}^t,~z_{n-1}^f$ \textbf{to} $z_{n-1}$
\State Softmax $z_{n-1}$ \textbf{to} $y^*_i$
\State $\omega^{t*},\delta^{t*},\omega^{s*},\delta^{s*}=argmin_{\omega,\delta}$Distance$(y^*_i,y_i)$
\State $\omega^t\leftarrow\omega^{t*},~\delta^t\leftarrow\delta^{t*},~\omega^s\leftarrow\omega^{s*},~\delta^s\leftarrow\delta^{s*}$
\EndWhile
\EndWhile
\State 
\Return $\omega^t,\delta^t$~(Output parameters of Temporal branch)$;~\omega^s,\delta^s$~(Output parameters of Spatial branch).
\end{algorithmic}
\end{algorithm}

\section{Experiments}
In this section, we select samples from patients according to different tasks. In next subsections, we will first introduce the evaluation metrics. Then, we present the results of our approach in the MIMIC-IV dataset and also discuss these experimental results. In particular, based on the results of the experiments, we invited Professional Doctors to analyze the data and give us useful clinical feedback. This showed that the model could be used and was effective.

\subsection{Evaluation Metrics}
Predicting in-hospital mortality is a binary classification task, so the area under the receiver operating characteristic (AUC-ROC) is the main way to measure how well it works, and the area under the Precision-Recall (AUC-PR) is the supplement. The main metric for length-of-stay prediction is Median Absolute Deviation (MAD) which is lower to indicate better performance, and Cohen’s linear weighted kappa score (KAPPA) that the higher the better is usually used as a supplement. Most patients have more than one condition, so it's clear that phenotype classification is a multi-label classification problem, and we chose macro- and micro-averaged AUC-ROC as the evaluation criteria for this problem. Predicting decompensation is made easier by treating it as a binary classification task like predicting death, which is also mainly judged by the AUC-ROC.

\subsection{Results}

\begin{table*}[]
\caption{Result on in-hospital mortality prediction.}
\label{tab:comparison}
\centering
\begin{tabular*}{0.8\hsize}{@{\extracolsep{\fill}}l l c c }
\toprule
&Model   & AUC-ROC &AUC-PR\\ 
\midrule  
\multirow{2}{*}{baseline}  &Logic Regression~\cite{DBLP:journals/corr/HarutyunyanKKG17}   & 0.848 & 0.474\\ 
\multirow{2}{*}{} 
&LSTM~\cite{DBLP:journals/corr/HarutyunyanKKG17}   & 0.855  & 0.485 \\  
\midrule
\multirow{10}{*}{others}  &channel-wise LSTM~\cite{DBLP:journals/corr/HarutyunyanKKG17}   & 0.862 & 0.515  \\ 
\multirow{7}{*}{} &SAnD~\cite{song2018attend}  &0.857 & 0.518 \\
\multirow{7}{*}{} &TimeNet~\cite{song2018attend}   & 0.764  & 0.813\\
\multirow{7}{*}{} &TPC~\cite{DBLP:journals/corr/abs-2007-09483}   & 0.905  &0.691\\ 
\multirow{7}{*}{} 
&Cox Time-Varying\cite{royalty2021machine}   & 0.740  & 0.290\\ 
\multirow{7}{*}{} 
&BoXHED\cite{royalty2021machine}   & 0.780  &0.350 \\  
\multirow{7}{*}{} 
&Bagging~\cite{pattalung2021comparison}   & 0.780  & \verb|\| \\ 
\multirow{7}{*}{} 
&Gradient boosting~\cite{pattalung2021comparison}   & 0.830  &  \verb|\|\\ 
\multirow{7}{*}{} 
&Random 
Forests~\cite{pattalung2021comparison}   & 0.820   &  \verb|\|\\
\multirow{7}{*}{} 
&GRU-D\cite{wang2020mimic}   & 0.876  &0.532 \\
\midrule

\multirow{2}{*}{TSCAN}  &Feature concatenation   & 0.859 $\pm$ 0.025 &0.491 $\pm$ 0.026\\ 
\multirow{2}{*}{}  & Max pooling  & \textbf{0.907 $\pm$ 0.020}  &\textbf{0.692 $\pm$ 0.019}\\ 
\bottomrule
\end{tabular*}
\end{table*}

\begin{table*}[!ht]
\caption{Result on length-of-stay prediction.}    
\label{tab:los_comparison}
\centering
\begin{tabular*}{0.8\hsize}{@{\extracolsep{\fill}}llcc}
\toprule
&Model   & KAPPA   &MAD \\ 
\midrule  
\multirow{2}{*}{baseline}  &Logic Regression~\cite{DBLP:journals/corr/HarutyunyanKKG17}   & 0.402  & 162.3  \\ 
\multirow{2}{*}{} 
&LSTM~\cite{DBLP:journals/corr/HarutyunyanKKG17}   & 0.438    & 123.1 \\  

\midrule
\multirow{2}{*}{others}  &channel-wise LSTM~\cite{DBLP:journals/corr/HarutyunyanKKG17}   &  0.442 &  136.6 \\ 
\multirow{5}{*}{} 
\multirow{5}{*}{} 
&SAnD~\cite{song2018attend}   & 0.429    & \verb|\| \\  

\midrule
TSCAN  &Max pooling   & \textbf{0.451 $\pm$ 0.013}   & \textbf{120.1 $\pm$ 1.3}  \\ 
\bottomrule
\end{tabular*}
\end{table*}

\begin{table*}[!ht]
\caption{Result on phenotype classification.}    
\label{tab:phen_comparison}
\centering
\begin{tabular*}{0.8\hsize}{@{\extracolsep{\fill}}llcc}
\toprule
   & Model   &\makecell[c]{Macro AUC-ROC} &\makecell[c]{Micro AUC-ROC} \\ 
\midrule
\multirow{2}{*}{baseline}  &Logic Regression~\cite{DBLP:journals/corr/HarutyunyanKKG17}   & 0.739 & 0.799 \\ 
\multirow{2}{*}{} 
&LSTM~\cite{DBLP:journals/corr/HarutyunyanKKG17}   & 0.770    & 0.821\\  
\midrule

\multirow{2}{*}{others}  & \makecell[l]{channel-wise LSTM}~\cite{DBLP:journals/corr/HarutyunyanKKG17}   & 0.776 & 0.825 \\ 
\multirow{2}{*}{} 
\multirow{2}{*}{} 
&SAnD~\cite{song2018attend}   & 0.766    & 0.816\\  
\multirow{2}{*}{} 
&TimeNet~\cite{song2018attend}   & 0.764  & 0.813\\

\midrule
TSCAN & Max pooling  & \textbf{0.795$ \pm$ 0.022}   & \textbf{0.839 $\pm$ 0.018} \\
\bottomrule
\end{tabular*}
\end{table*}

\begin{table*}[!ht]
\caption{Result on decompensation prediction.}    
\label{tab:decom_comparison}
\centering
\begin{tabular*}{0.8\hsize}{@{\extracolsep{\fill}}llcc}
\toprule
 &Model   & AUC-ROC    & AUC-PR \\ 
\midrule
\multirow{2}{*}{baseline}  & Logic Regression~\cite{DBLP:journals/corr/HarutyunyanKKG17}   & 0.870 & 0.214 \\ 
\multirow{2}{*}{} 
&LSTM~\cite{DBLP:journals/corr/HarutyunyanKKG17}   & 0.892    & 0.324\\ 
\midrule

\multirow{5}{*}{others}  &channel-wise LSTM~\cite{DBLP:journals/corr/HarutyunyanKKG17}   & 0.906 & 0.333 \\ 
\multirow{5}{*}{} 
\multirow{5}{*}{} 
&SAnD~\cite{song2018attend}   & 0.895    & 0.316\\ 
\multirow{5}{*}{} 
&CNN-RNN~\cite{10.1145/3219819.3220051}   & 0.874    & 0.231\\  
\multirow{5}{*}{} 
&CNN-AttRNN~\cite{10.1145/3219819.3220051}   & 0.881    & 0.258\\  
\multirow{5}{*}{} 
&RAIM~\cite{10.1145/3219819.3220051}   & 0.901    & 0.279\\ 
\midrule
TSACN  &Max pooling   & \textbf{0.913 $\pm$ 0.021}  & \textbf{0.326 $\pm$ 0.019} \\ 
\bottomrule
\end{tabular*}
\end{table*}

Traditionally, Logistic Regression (LR) was the most accurate model for medical time series. In recent years, LSTM and some deeplearning methods have done well with time series. In this section, Logistic Regression~\cite{DBLP:journals/corr/HarutyunyanKKG17} and LSTM~\cite{DBLP:journals/corr/HarutyunyanKKG17} were used as a baseline and compared with our approach. 

For in-hospital mortality prediction, the prediction time point is defined as 48h after the beginning of the ICU stay, and $t$=48h which means one ICU stay is one sample. As can be seen from the Table.\ref{tab:comparison}, TSCAN shows a huge improvement in classification scores relative to the baseline, with feature concatenation improving by 0.011 relative to LR and Max pooling improving by 0.052 relative to LSTM, and we have achieved significant performance benefits of 0.2\% compared to other SOTA methods (TPC) in ACU-ROC.

For length-of-stay prediction, we select prediction time point every 12 hours, begining at the fourth hour after the patient is admitted to the ICU and ending with the patient's discharge or death. $t=320$h, $n$=4 and the task aims to predict the remaining time spent of patients in ICU. As shown in Table.\ref{tab:los_comparison}, TSCAN realize a reduction of three points compared to LSTM and more than forty points compared to logistic regression in MAD. For the metric of KAPPA, TSCAN also improves by 0.009 compared to channel-wise LSTM.

For phenotype classification, $n=4$, the prediction time point is described as the end of the ICU stay (the patient is discharged or dies), and the time-series data of $t=320$h are used to classify phenotype in order to make full use of the data during the patient's stay in the ICU. In other words, for phenotype classification, those less than 320 in length in the temporal dimension are padded with zero, and those that are redundant are removed. As shown in the Table.~\ref{tab:phen_comparison}, it can be seen that TSCAN improves by 0.056 in macro-averaged AUC-ROC and 0.040 in micro-averaged AUC-ROC over logistic regression. Compared to LSTM, the experiment shows an improvement of 0.025 in the macro-averaged AUC-ROC and 0.018 in the micro-averaged AUC-ROC. This shows that the internal Fusion-Encoder module also works well for problems with more than one label.

For decompensation prediction, $n=4$, we select a prediction time point every hour, 4h after the beginning of the ICU stay, and $t=320$h. And the results are shown in the Table~\ref{tab:decom_comparison}. It can be seen that TSCAN has an improvement of 0.043 in the AUC-ROC metric and 0.112 in the AUC-PR metric compared to Logic Regression, and 0.021 in the AUC-ROC metric compared to LSTM. And we have achieved significant performance benefits of 0.7\% compared to other SOTA methods (channe-wise LSTM) in ACU-ROC. Even when the data is pre-processed to get a time-equal sequence, it is clear that our method can still do a great job.

\subsection{Discussion on In-hospital Mortality}
We take data from the initial 48 hours of ICU stays as the sample for the prediction task. Early prediction of hospital deaths can help identify patients who are at high risk and give medical staff an alert. In the next subsections, we apply this task to define some parameters of a prediction model that will be utilized in other tasks. 

\subsubsection{Number of Indicators}
\begin{figure}[]
\centering
\includegraphics[width=0.45\textwidth]{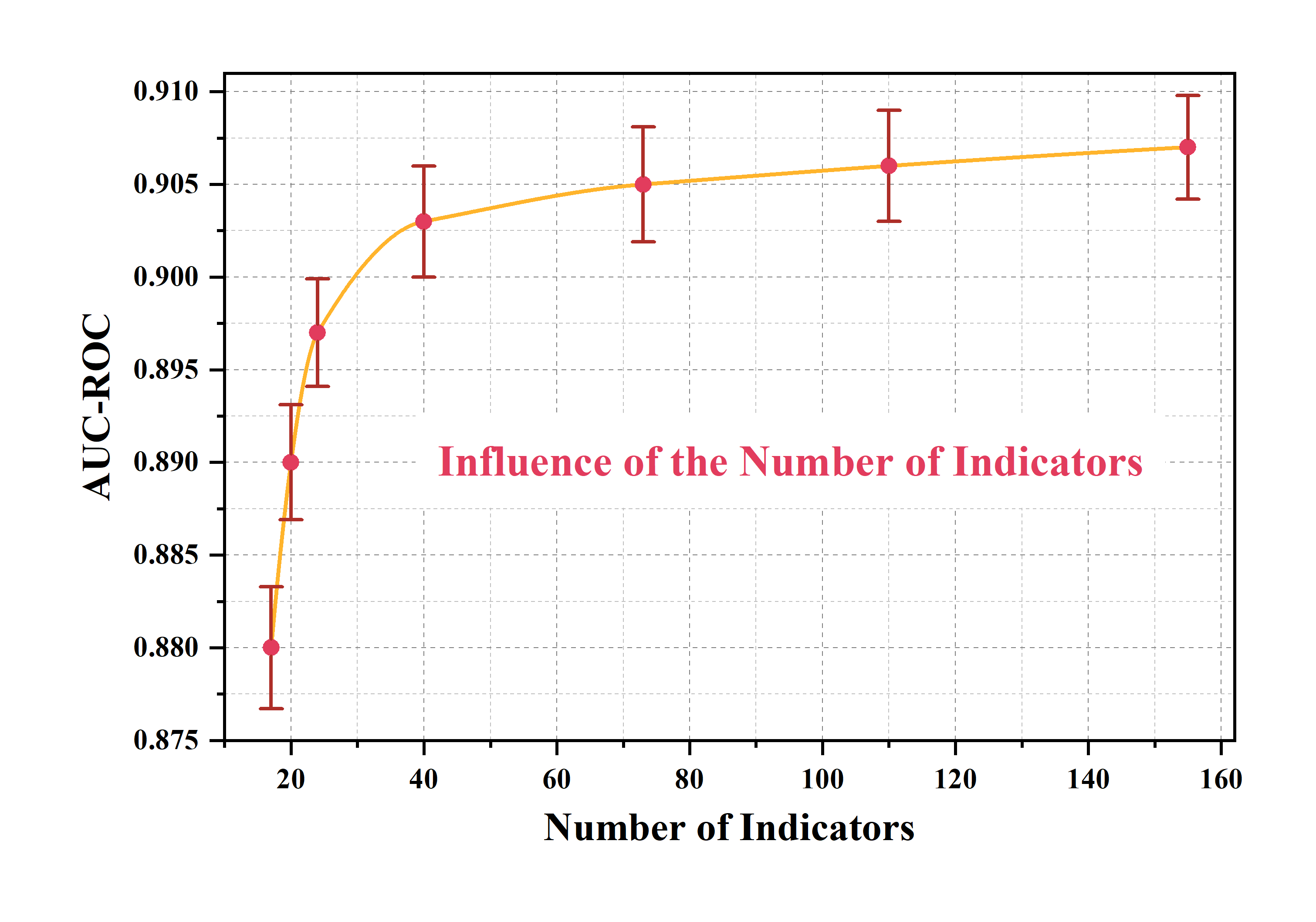}
\caption{\label{fig:indicators}Results of different numbers of indicators on in-hospital mortality prediction. With the number of indicators increasing from 17 to 155, the growth rate of AUC-ROC is levelling off.}
\end{figure}

In this paper, data from 155 clinical indicators was selected to handle related prediction tasks based on the recommendations of the local hospital. From a medical point of view, the accuracy of mortality prediction with different number of variables as reference is expected to vary, theoretically setting the higher the number of variables the higher the accuracy. To compare the effect of the number of variables on the results of the experiment, we randomly select clinical data from 17-155 to verify the impact of variables on prediction accuracy. The related experimental results are shown in Fig.\ref{fig:indicators}.

As shown in Fig.\ref{fig:indicators}, The number of indicators increases from 17 to 155, demonstrating that the accuracy of mortality prediction improves with the number of variables, which is congruent with objective evidence. It is shown that our model can well exploit the indicators suggested by the hospital and make full use of them in the prediction task. However, we also find that the rate of growth is gradually decreasing. This condition means that continued increases in the number of indicators may not achieve a significant improvement in forecasting. Due to data limitations, the following task also utilize 155 variables as input.

\subsubsection{Ablation Studies}
As shown in Fig.\ref{fig:model}, the model of the scheme can be separated into two branches: the temporal domain and the characteristic domain (i.e. the spatial domain), so the ablation experiments are mainly based on the comparison of these two branches and the comparison of different fusion methods. We conducted experiments in the spatial and temporal domains respectively, and then fused the two branches in different ways. Four fusion methods are selected including feature concatenation, adding fusion, bilinear pooling and max pooling (prediction followed by fusion). The results of the ablation studies are shown in the Table.\ref{tab:ihm_ablation}.
\begin{table*}[!htbp] 
\caption{Ablation studies on in-hospital mortality prediction.}    
\label{tab:ihm_ablation}
\centering
\begin{tabular*}{\hsize}{@{\extracolsep{\fill}}lcccccc}
\toprule
studies     &\makecell[c]{Temporal\\domain}     &\makecell[c]{Spatial\\domain}     &\makecell[c]{concatenate fusion}     &\makecell[c]{Adding fusion}     &\makecell[c]{Bilinear pooling~\cite{DBLP:journals/corr/LinRM15a}}     &\makecell[c]{Max Pooling}     \\
\midrule  
AUC-ROC     &0.857     &0.832     &0.859     &0.856     &0.843     &\textbf{0.907}  \\
\bottomrule
\end{tabular*}
\end{table*}

As seen in the table above, the performance of attention in the temporal domain alone is superior to that in the spatial domain alone, with an AUC-ROC score of 0.025 greater the former than the latter . Analyzing the sources of this conclusion reveals that, for medical time series, the correlation between adjacent times is extremely high, although the independence between variables is relatively high. 
It is also obvious that combining the temporal and spatial dimensions gives better metrics, with max pooling fusion providing the best results and concatenate fusion being the next best.

\subsubsection{Attention in Temporal Domain}
\begin{figure}[t]
\centering
\includegraphics[width=0.5\textwidth]{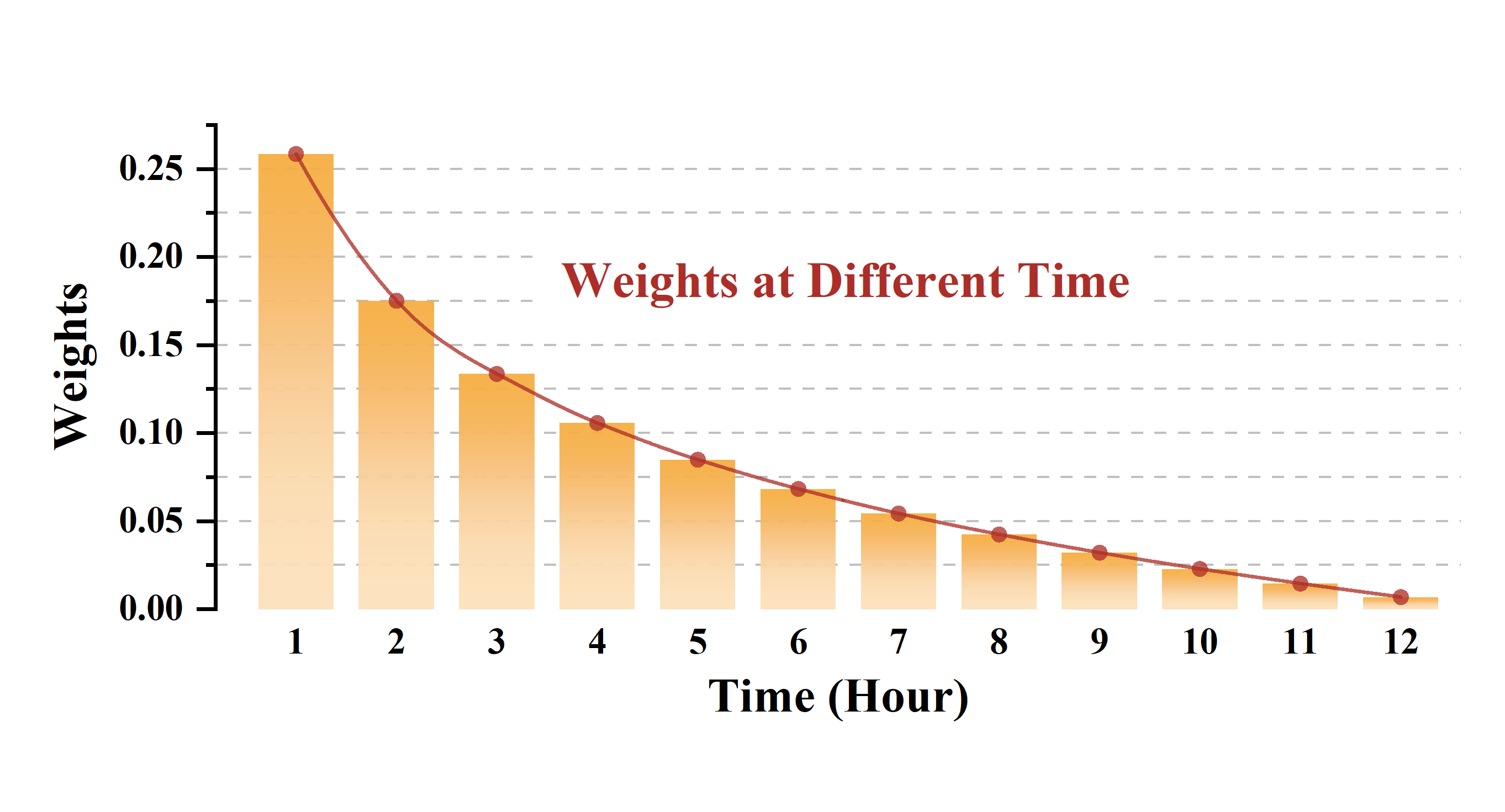}
\caption{\label{fig:ihm_wt} Weights of times on in-hospital mortality prediction. Each encoder takes 12 hours of data input, corresponding to the twelve scales on the horizontal axis in turn, and the vertical axis shows the weights that each hour occupies.}
\end{figure}

In this paper, we apply sequence time data to handle prediction problem. Naturally, We hope to find the contribution of data at different time points to the prediction results. From in-hospital mortality prediction task, we divide 48 hours' data into 4 parts and each part include 12 hours' data. We can obtain 4 attention maps. Here, we  calculate the weights and visualize it like Fig.\ref{fig:ihm_wt}. The horizontal axis represents the data of 12 moments and the vertical axis represents the correlation extracted of the temporal dimension.

From these visual result, we can find that the closer to the prediction time point, the more contribution to the prediction. As the distance from the predicted time point increases, the correlation gradually decreases. This is also combined with the actual clinical diagnosis and treatment situation. We also modified the parameter $n$ of the data grouping and got similar results. 
For other prediction and classification tasks, we achieve the similar results. 
Due to the length of the paper, the other tasks are not to discuss the weight of data in time. 

\subsubsection{Attention in Clinical Indicators}
Attention mechanism is used to get the weights of each indicator. We hope to find out how each of the 155 chosen indicators helps with the task. Attention extraction in the spatial domain can help us reach this goal.
\begin{figure}[t]
\centering
\includegraphics[width=0.4\textwidth]{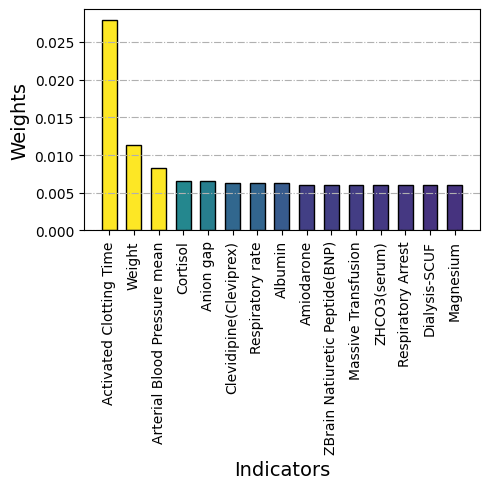}
\caption{\label{fig:ihm_wf} Indicator weights for predicting in-hospital mortality The horizontal axis shows the 15 variables with the highest weighting and the vertical axis indicates their contribution to the prediction of in-hospital mortality.}
\end{figure}

As shown in Fig.\ref{fig:ihm_wf}, the horizontal axis shows the 15 indicators with the highest contribution ratios of the 155 used in the experiment. The vertical axis shows how much each variable contributed to the prediction of in-hospital mortality. For example, Activated Clotting Time made up more than 0.025 of the weights, which is the highest percentage. Knowing how important each indicator is can help doctors decide what to focus on and how to treat their patients. 
These results show that coagulation is a major factor in a patient's chance of survival, as is the use of Amiodarone for arrhythmias, SCUF for heart failure, and Magnesium for treating patients with disorders of the body's internal environment. The above situation has been affirmed by professional doctors, and it also proves that our approach has obvious advantages in the discovery of the cause of the disease, and can be applied to similar prediction tasks and the discovery of potential causes of the disease, such as DNA screening and functional discovery.

\subsection{Discussion on Length-of-stay Prediction}
This task is to predict remaining time spent in ICU at every 12 hours of stay, which is framed as a classification problem with 10 buckets (one for ICU stays shorter than a day, seven day-long buckets for each day of the first week, one for stays of over one week but less than two, and one for stays of over two weeks). With this work, ICU beds can be used more efficiently and patients can be cared for better.
\begin{figure}[t]
\centering
\includegraphics[width=0.4\textwidth]{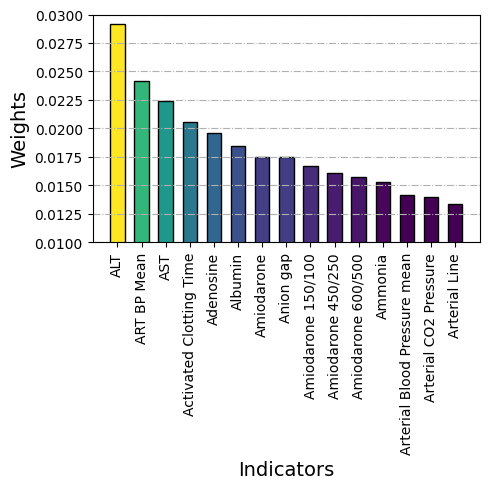}
\caption{\label{fig:los_wf} Indicator weights for predicting length of stay It can be seen that Activated Clotting Time accounts for the highest percentage of weights at near 0.03.}
\end{figure}

We also focus on the contribution of the selected 155 indicators to the task. As shown in Fig.\ref{fig:los_wf}, the horizontal axis shows the 15 variables with the highest proportion of the 155 variables used in the experiment, and the vertical axis indicates their contribution to the prediction of length-of-stay.
From these results, we can see that ALT is responsible for the most weights, which is close to 0.03.


\subsection{Discussion on Phenotype Classification}
Phenotyping has applications in cohort construction for clinical studies, comorbidity detection and risk adjustment, quality improvement and surveillance, and diagnosis~\cite{agarwal2016learning}. Based on the ICU data, the phenotyping classification task is to figure out which type of acute care the patient is in. There are 25 common conditions selected for this paper, including 8 chronic conditions that are common co-morbidities and risk factors in ICU such as essential hypertension, 12 severe conditions that are relatively more dangerous to life such as pneumonia, and 5 mixed conditions that are recurrent or chronic, with periodic acute episodes, as shown in Table.\ref{tab:25_phenotype}.

\begin{table}[t]
\caption{ICU phenotypes. There are 25 common conditions selected for this paper including 8 chronic conditions, 12 severe conditions and 5 mixed conditions.}    
\label{tab:25_phenotype}
\begin{tabular*}{\hsize}{@{\extracolsep{\fill}}ll}
\toprule
Phenotype                           & Type            \\
\midrule
Acute and unspecified renal failure  & severe \\
Acute cerebrovascular disease   & severe \\
Acute myocardial infarction & severe  \\
Cardiac dysrhythmias & mixed   \\
Chronic kidney disease  & chronic   \\
Chronic obstructive pulmonary disease  & chronic \\
Complications of surgical/medical care  & severe \\
Conduction disorders & mixed \\
Congestive heart failure; nonhypertensive & mixed \\
Coronary atherosclerosis and related  & severe  \\
Diabetes mellitus with complications  & mixed  \\
Diabetes mellitus without complication & severe  \\
Disorders of lipid metabolism & severe  \\
Essential hypertension & severe  \\
Fluid and electrolyte disorders & severe  \\
Gastrointestinal hemorrhage & severe  \\
Hypertension with complications & severe  \\
Other liver diseases & mixed  \\
Other lower respiratory disease & severe  \\
Other upper respiratory disease & severe  \\
Pleurisy; pneumothorax; pulmonary collapse & severe  \\
Pneumonia & severe  \\
Respiratory failure; insufficiency; arrest & severe  \\
Septicemia (except in labor) & severe  \\
Shock & severe  \\
\bottomrule
\end{tabular*}
\end{table}

\begin{figure}[t]
\centering
\includegraphics[width=0.4\textwidth]{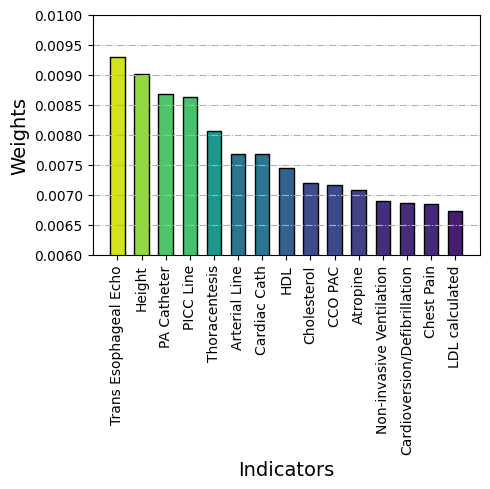}
\caption{\label{fig:phen_wf} Weights of indicators on phenotype classification. The horizontal axis presents the top 15 variables in terms of proportion and the vertical axis indicates their contribution to the task.}
\end{figure}

The attention map of clinical indicators is shown in Fig.\ref{fig:phen_wf}. The weights of the top 15 variables are shown on the horizontal axis, and their contributions to the task of phenotyping classification are shown on the vertical axis. Several indicators such as the esophageal echo, height, PA catheter, and PICC line accounted for a higher weighting of over 0.008, which suggests that medical staff could pay more attention to those factors when judging the phenotype.


\subsection{Discussion on Decompensation Prediction}
The goal of this task is to figure out which patients are likely to get much worse in the next 24 hours. This setting is currently manually designed with early warning scores and thresholds below which alerts are triggered, so most of these scoring systems are based on simple thresholds and a small number of common physiological indicators, such as the National Early Warning Score (NEWS)~\cite{rcopo2012national} and the Modified Early Warning Score (MEWS)~\cite{subbe2001validation}. Decompensation prediction in the paper aims to detect patients who are physiologically decompensating, or whose conditions are deteriorating rapidly.

Referring to previous work~\cite{DBLP:journals/jbi/PurushothamMCL18}, and trying to fit the early warning scoring system as closely as possible, the task of decompensation prediction in our experiment is considered as a prediction of mortality in the next 24 hours for patients in the ICU and is done every hour. So, our experiments can be run on the MIMIC-IV database, which will pull out correct features and labels.

For each hour as a prediction time point, the data is matched to a target label that says whether the patient died within 24 hours of that hour. In contrast to predicting in-hospital mortality and phenotype classification, for a single ICU stay, more than one sample can be made.

\begin{figure}[t]
\centering
\includegraphics[width=0.4\textwidth]{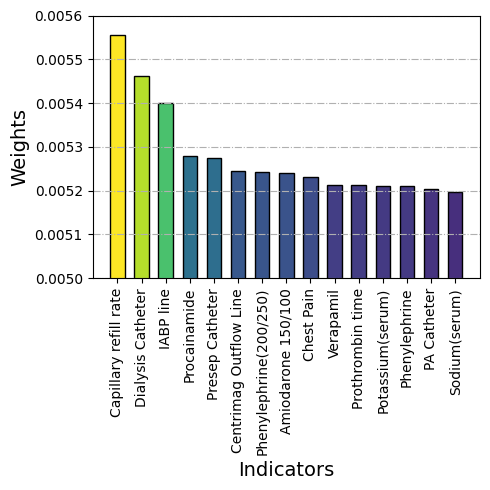}
\caption{\label{fig:decom_wf} Weights of indicators on decompensation prediction. The vertical axis shows the contribution ratios of the first 15 variables, which account for the most weight. Indicators with the highest weight is Capilary refill rate.}
\end{figure}

The attention map of clinical indicators is shown in Fig.\ref{fig:decom_wf}. The vertical axis shows the contribution ratios of the first 15 variables, which account for the highest weight in the task. The percentages for the top 15 variables are relatively average, ranging from 0.005 to 0.006. The predictions are consistent with clinical reality. For the PICC line, the patient needs intravenous nutrition and rehydration therapy, and for the PA catheter, the patient needs to be checked for complex hemodynamic disturbances. Both of these are treatments for patients who are pretty sick.

On the one hand, the results suggest that patients with IABP on left heart assist devices, arrhythmia drugs such as amiodarone, verapamil, and chest pain might be more likely to deteriorate, while patients with cardiac dysfunction might be less likely to survive. On the other hand, the capillary refill rate and phenylephrine are important predictors of a patient's prognosis in cases of circulatory failure. This means that stable circulation and enough blood flow to tissues and organs are needed to improve the prognosis of a patient. The above indicates that cardiovascular system disease plays an important role in predicting decompensation.

\section{Conclusion}
The widespread use of EHR and the fast growth of deep learning have made it possible for medical time series to be processed and used in new ways. In this paper, we use the MIMIC-IV database to build a deep learning model for medical time series. This model does well at predicting in-hospital mortality, with evaluation metrics that are much better than the baseline. First, based on an attention mechanism,
the segmentation of the data and internal structure of the Fusion-Encoder in TSCAN give full consideration to temporal correlation of the clinical time series. Secondly, our model can also find key clinical indicators of important outcomes that can be used to improve treatment options. This has an effective guiding role for clinical diagnosis and treatment in practice.

\section*{Acknowledgment}

This work was supported in part by the National Natural Science Foundation of China (62272337, 61902277) and the Natural Science Foundation of Tianjin (16JCZDJC31100).

\bibliographystyle{bibtex/IEEEtran}
\bibliography{bibtex/reference}





\end{document}